\documentclass{article} 
\usepackage{nips12submit_e,times}

\usepackage{booktabs} 
\usepackage{array} 
\usepackage{paralist} 
\usepackage{verbatim} 
\usepackage{subfig} 
\usepackage{todonotes}
\usepackage{mathtools}

\usepackage[numbers]{natbib}

\title{Compiling Relational Database Schemata into Probabilistic Graphical Models}

\author{
Sameer Singh\\
University of Massachusetts, \\
Amherst MA, USA \\
\texttt{sameer@cs.umass.edu}
\And
Thore Graepel \\
Microsoft Research, \\
Cambridge, United Kingdom \\
\texttt{thoreg@microsoft.com}
}

\newcommand{\cut}[1]{}

\renewcommand{\subsection}[1]{\textbf{#1:}}

\nipsfinalcopy 

\begin{document}

\maketitle

\section{Introduction}
\label{sec:intro}

A majority of scientific and commercial data is stored in relational databases.
Probabilistic models over such datasets would allow probabilistic queries, error checking, and inference of missing values, but to this day machine learning expertise is required to construct accurate models.
Fortunately, current probabilistic programming tools ease the task of constructing such models~\citep{pfeffer01:ibal:,pfeffer09:figaro:,milch06:probabilistic,goodman08:church:,mccallum09:factorie:,minka10:infernet} and work in statistical relational learning has focused on making it even easier to define models specific to relational data~\citep{friedman99:prm,taskar02:discriminative,heckerman04:daper,neville07:relational}.
However, within these frameworks the user still needs to specify all the probabilistic dependencies in the data, requiring a level of expertise in probability and statistics that domain experts often do not have, thus severely restricting the practical applications of such techniques.
On the other hand, domain experts do spend considerable effort and expertise in designing the database schemata used to represent their data, providing type information for table columns and foreign key relations to specify dependencies.

In this work, we view relational database schemata as \emph{programs} that describe probabilistic dependencies that exist in the data. The goal is to simplify the task of model construction for the domain expert and to be able to construct probabilistic models automatically for a large number of existing databases without manual intervention.
Using a given schema, a customized fully-Bayesian, generative graphical model is generated. Each table is modeled with a mixture model, along with edges that model dependencies between these table models according to their foreign key relationships.
This underlying model is similar to relational latent variable models~\citep{xu06:infinite,kemp06:learning}, but extends them by incorporating referential uncertainty (foreign key prediction) and using a parametric approach for real-world tractability.
We use variational message passing inference to learn the parameters of the model, allowing inference of missing values and probabilistic relational queries.
Experiments demonstrate the accuracy and scalability of the approach using synthetic and real world data.


\section{Compiling a Graphical Model from the Schema}
\label{sec:model}

In this section, we describe how, given a database schema, we create a Bayesian graphical model and perform inference with minimal manual intervention.

\subsection{Single Table}
We begin the description of the model by examining a schema that contains a single table $A$ with attributes $\mathbf{x}^A$.
We employ a \emph{mixture model} for each table, wherein a mixture component is used to generate all the attributes $\mathbf{x}^A(i)$ of row $i$, and $z^A(i)$ is a latent variable that indicates which component to use for the row.
The distribution used to generate each attribute $x_k^A(i)$ depends on the data type of the attribute; Gaussian for real-valued, Discrete for categorical-valued, and Bernoulli for Boolean-valued attributes, each distribution is latent and generated from its observed prior.
The component indicators $\mathbf{z}^A$ are generated from a latent discrete distribution $\pi^A$, with its observed prior.

\subsection{Foreign Component Link}
Consider a table $B$ that contains a single foreign key attribute to another table $A$.
The data attributes of both tables $A$ and $B$, $\mathbf{x}^A$ and $\mathbf{x}^B$ respectively, are modeled as described above.
The foreign key attribute for each row $i$ in table $B$ is represented by $f^B_i$, which indexes into a row in table $A$.
Since we want the links between rows to reflect the dependencies between the tables, we make the component indicator $z^B(i)$ dependent on the component indicator of the foreign row it links to ($z^A(f^B_i)$).
Specifically, instead of using a single distribution for $z^B$, we use as many discrete distributions as the number of components in table $A$ (cardinality of $z^A$), and \emph{select} the corresponding distribution using Gates~\citep{minka08:gates}: $z^A(i)\leftarrow \pi^B[z^A(f^B_i)]$.
We also model the uncertainty in foreign keys $f^B$ as discrete distributions, which allows prediction of missing foreign links.
This idea is easily generalized to tables with an arbitrary number of foreign keys by using additional number of discrete distributions $\pi^B$.

\subsection{Database Schema}
\begin{figure}[tb]
\begin{center} \subfloat[Schema]{\raisebox{10mm}{\includegraphics[width=0.2\textwidth]{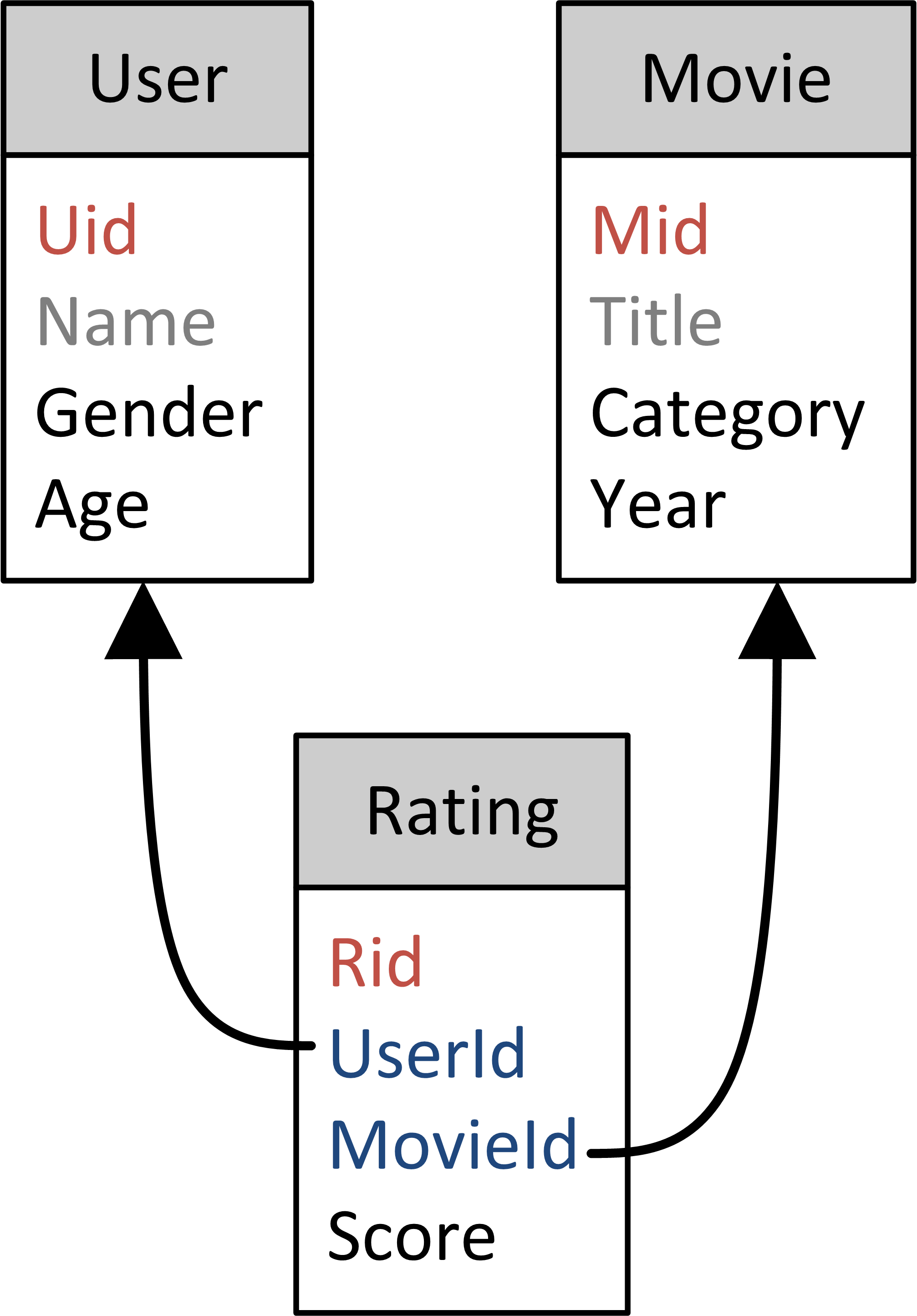}}\label{fig:umexample:schema}}\qquad
  \subfloat[Generated Graphical Model]{\includegraphics[width=0.4\textwidth]{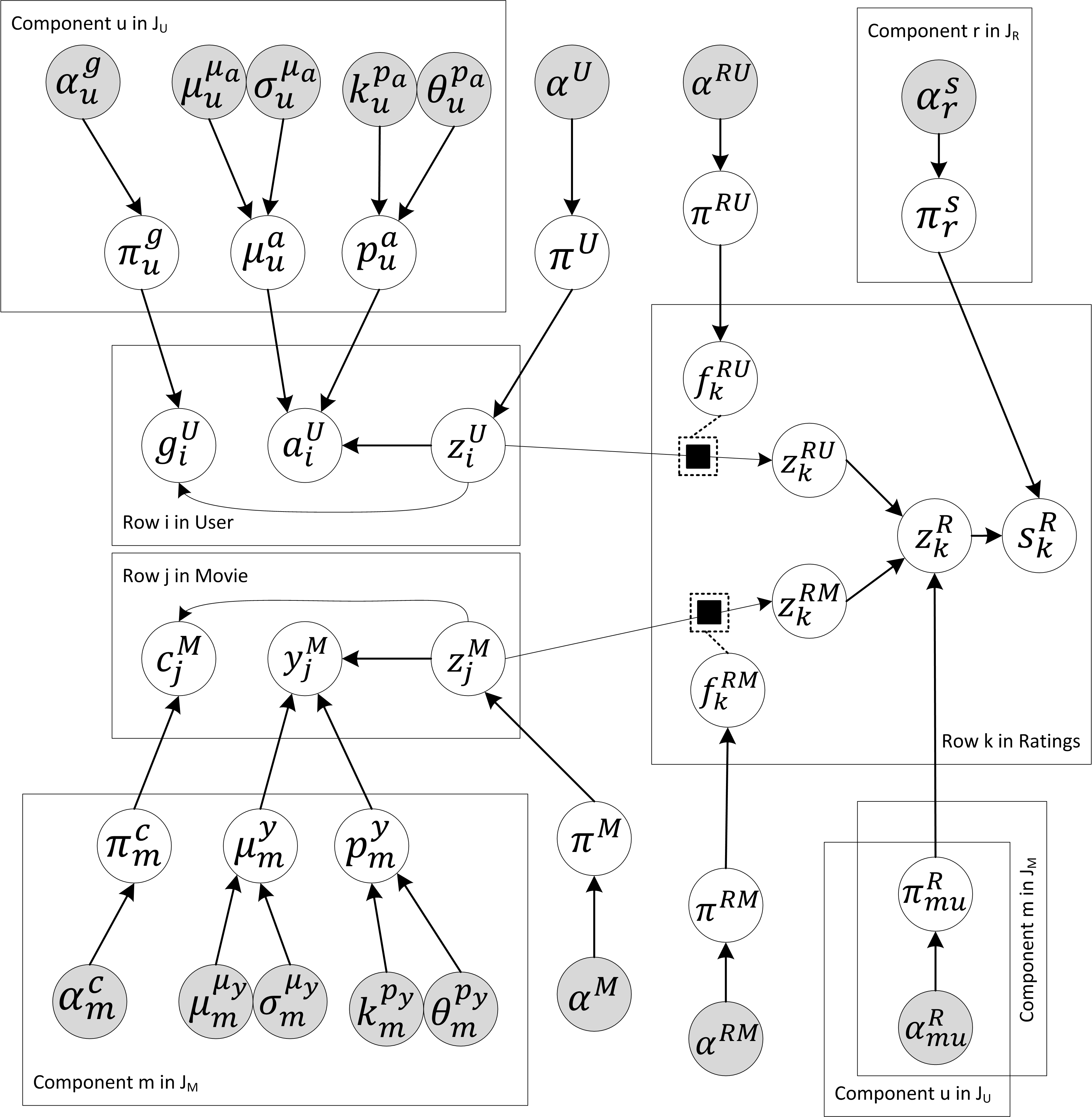}\label{fig:umexample:model}}
  \caption{{\small{\bf User-Movie-Rating Schema:} Example schema (a) consisting of movie ratings by users. Attributes shown in gray and primary keys (red) are not modeled. The data attributes (black) are represented by variables $g^U$, $a^U$, $c^M$, $y^M$, $s^R$ in the model (b). The foreign key relations (blue) are modeled using $f_k^{RM}$ and $f_k^{RU}$.}}
  \label{fig:umexample}
\end{center}
\end{figure}

An input database schema consists of a number of tables and their attributes, and the foreign key relations that form a directed, acyclic graph.
We can use the building blocks above to iteratively construct a model over a schema by applying the single-table model for the tables without any foreign keys, and using the foreign links to define the dependencies between the component indicators for tables with foreign keys.
For example, consider a simple schema consisting of three tables shown in Figure~\ref{fig:umexample:schema}.
Figure~\ref{fig:umexample:model} shows the generated model, where the model for User and Movie tables is similar to a regular mixture model, while the Rating table consists of additional variables and edges for foreign links, and dependencies of the component indicators across tables.

\subsection{Model Assumptions}
As described, a number of priors in the models need to be specified.
Most hyper parameters can be set to be \emph{uninformative}, however specifying the number of components in each table is crucial.
Too many components result in slower inference, while too few components produce inaccurate models.
Non-parametric approaches such as \cite{xu06:infinite,kemp06:learning} are much slower in practice, however recent work suggests that exploiting conditional exchangeable properties of our data may be useful~\cite{lloyd12:random}.
Another assumption in the generated model is that the attributes of the row are independently generated given the component, which often does not hold in practice.
An alternative is to explore the range of independent to fully-correlated attributes, using cross-cutting models~\cite{shafto06:learning}.

\subsection{Inference}
Inference on the resulting model is performed using variational message passing~\cite{winn03:vmp}, as implemented in Infer.NET~\citep{minka10:infernet}.
Since the model contains strong dependencies and deterministic factors (\emph{gates}), inference approaches such as Gibbs sampling are not practical when applied directly.
During training, in which we learn the parameters of the model and use it to predict missing values in the database, the complexity of message passing is linear in the number of rows (when all the foreign keys are observed).
The approach also supports \emph{probabilistic queries} over the trained model; queries take the form of a small set of records with missing entries. Inference is used to predict marginal posterior belief distributions over these entries.
The inference for querying is also efficient; linear in the size of the query if the foreign keys are observed.

\section{Experiments}
\label{sec:exps}

In this section, we present preliminary experiments that evaluate the accuracy, clustering quality, and the scalability of the schema-based probabilistic models.

\begin{figure}[tb]
\begin{center}
  \subfloat[Rating Score of a Movie]{\label{fig:results:score}\includegraphics[trim = 6mm 5mm 3mm 6mm, clip, width=0.3\textwidth]{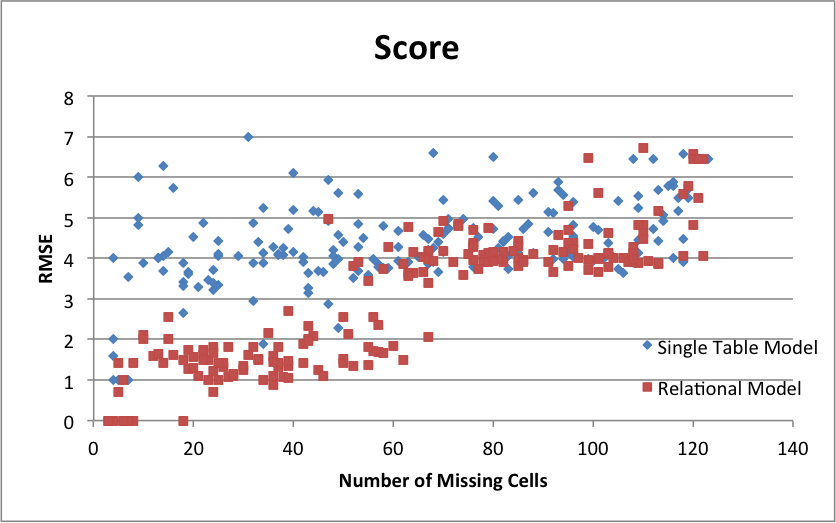}}\quad
  \subfloat[Age of a User]{\label{fig:results:age}\includegraphics[trim = 6mm 5mm 4mm 6mm, clip, width=0.3\textwidth]{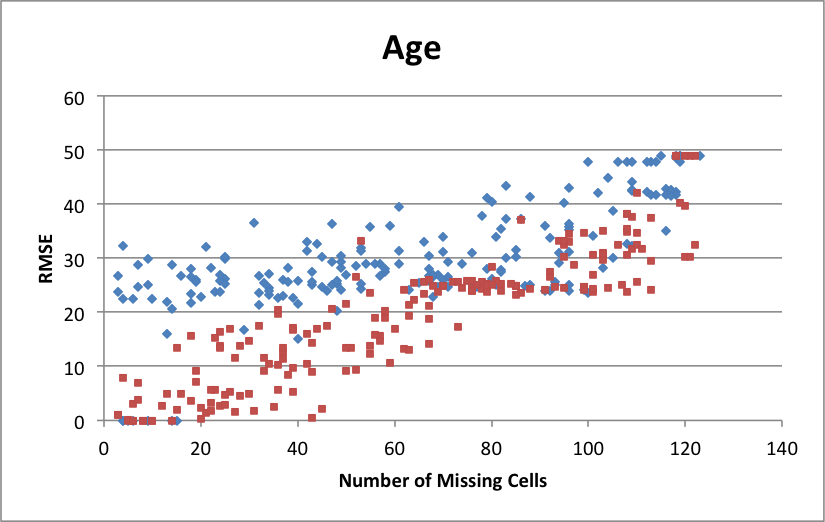}}\quad
  \subfloat[Release year of a Movie]{\label{fig:results:year}\includegraphics[trim = 6mm 5mm 4mm 6mm, clip, width=0.3\textwidth]{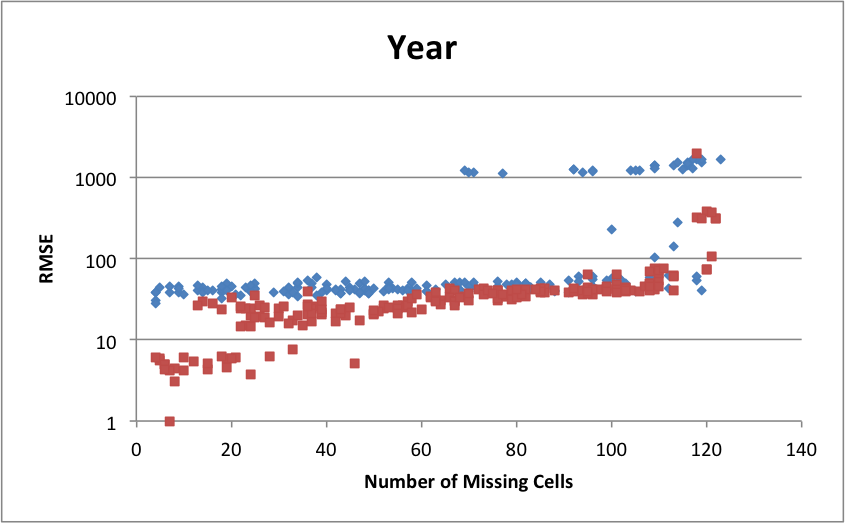}}
  \caption{\small{\bf Results on Synthetic Data:} Comparison of the relational model (\emph{in red}) with a single table model generated using a join over the foreign keys (\emph{in blue}) using RMSE on three real-valued attributes.}
  \label{fig:results:um}
\end{center}
\end{figure}

\subsection{Synthetic User-Movies-Ratings Data}
One typical approach to modeling values in a relational database is to perform a join over all the tables, and to use a single-table mixture model on the resulting table. 
Unlike in our relational model, the dependencies across rows are lost in the join operation.
To evaluate this effect on accuracy, we compare the two models by treating a proportion of cells as missing (before performing the join).
We create synthetic data for the schema in Figure~\ref{fig:umexample:schema}, and perform inference to predict the values of the missing cells.
The error of the predictions for the real-valued attributes is shown in Figure~\ref{fig:results:um}, demonstrating that the schema-based probabilistic model is consistently more accurate and more robust in the presence of missing cells.
In particular, the rating scores are accurate even when half of the values are missing.

\begin{figure}[tb]
\begin{center}
  \subfloat[Varying the number of rows] {\label{fig:results:ml}\includegraphics[width=0.35\textwidth]{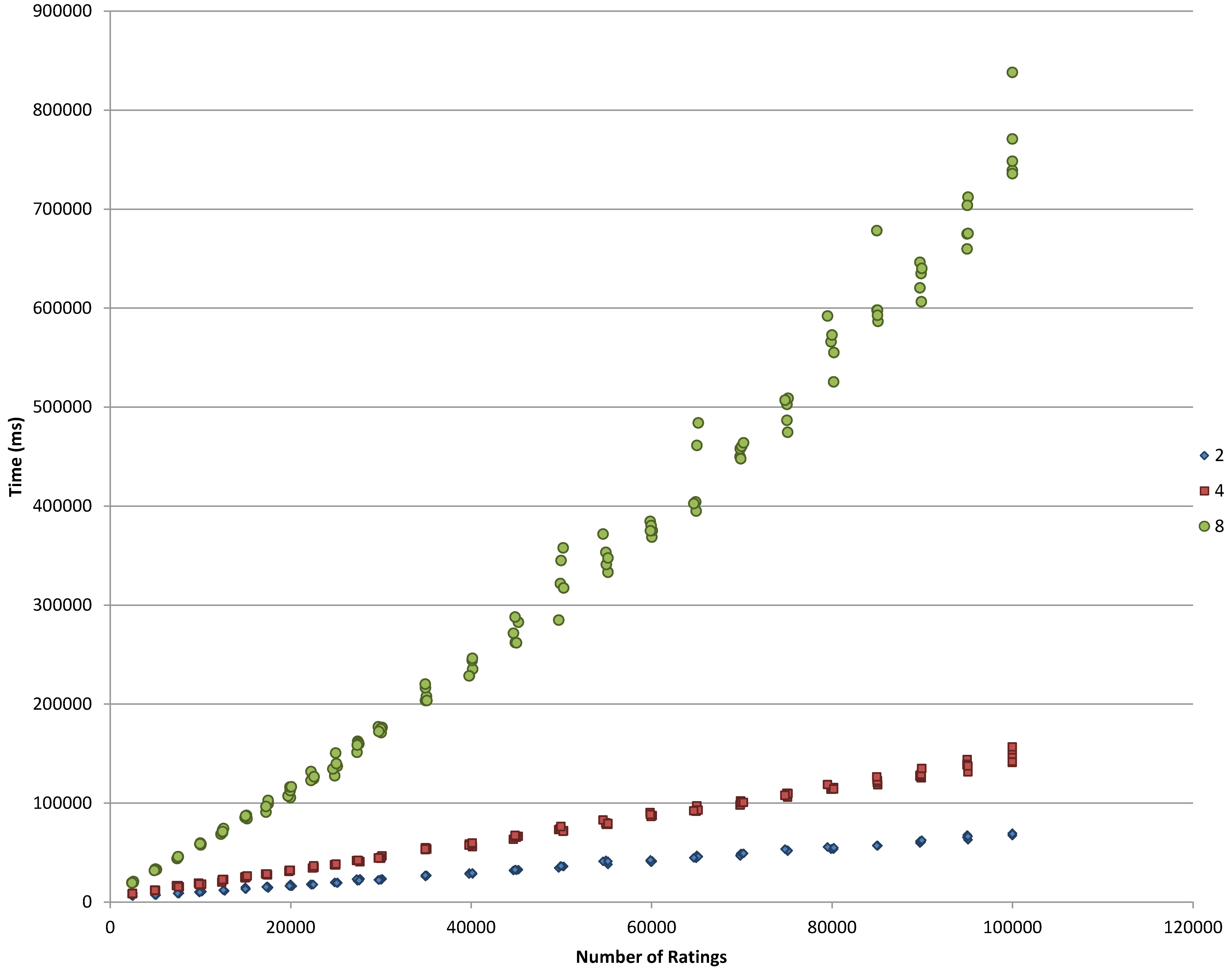}}\qquad
  \subfloat[Clustering of Xbox Players]{\label{fig:results:trueskill}\raisebox{3mm}{\includegraphics[width=0.3\textwidth]{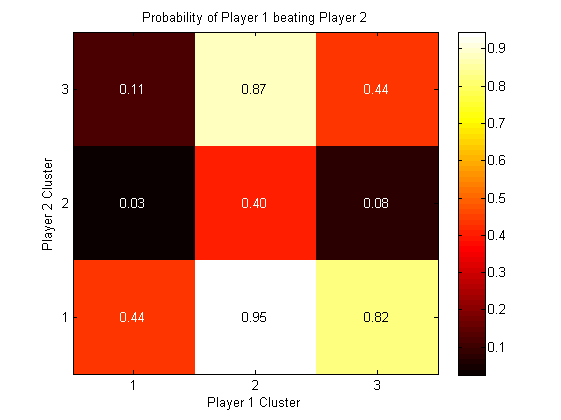}}}%
  \caption{\small{\bf Experiments on Real-World Data:} (a) MovieLens 100k, and (b) Xbox Head-on-Head data}
  \label{fig:results}
\end{center}
\end{figure}

\subsection{MovieLens dataset}
We evaluate scalability on the MovieLens dataset. 
The schema of the data is similar to User-Movie-Rating database, but includes a few more attributes.
The data consists of $943$ users, $1,682$ movies, and $100,000$ ratings.
Since the number of rows in the \emph{leaf} table is usually much higher than in other tables, we examine the scalability in terms of its size.
We run a fixed number of iterations of inference as we vary the number of ratings, and examine the running time.
The results, shown in Figure~\ref{fig:results:ml}, show a linear trend for the running time.
Further, the figure also shows the increase in running time as the number of components in each table is increased.

\subsection{TrueSkill Dataset}
To perform a qualitative evaluation of the clustering of rows produced by our model, we use the Head-to-Head games data from Xbox matches, as used in \citet{herbrich07:trueskill}.
The data consists of a table of player Ids (with no other attributes), and a table of match results that consists of foreign key attributes for two players, along with a Boolean result attribute that is true if the first player was the winner.
The model generated for this data assigns each player row to one of three components, shown in Figure~\ref{fig:results:trueskill}.
We also include the average result for each pair of clusters.
Note that the three clusters correspond to bad, excellent, and good players respectively, demonstrating that the latent clustering can be used to predict the skills of players without making any further domain-specific modeling assumptions. 

\section{Conclusion and Future Work}
\label{sec:conclusions}

We suggest automatically compiling probabilistic graphical models from database schemata. 
This approach allows us to make use of the domain knowledge that went into the design of the database schema and potentially makes probabilistic graphical models directly available for a large fraction of the world's data.
Inference on the compiled Bayesian model allows the prediction of the values of missing cells in the database, detect outliers, visualize clustering of the data, and to answer basic probabilistic relational queries.
We evaluated the accuracy, the clustering quality, and the scalability of our approach using a combination of synthetic and real world data, and found that the schema-based graphical models lead to interesting results.

This work is very much in progress, and there are a number of avenues for future directions.
We would like to explore computationally efficient extensions to the model that are non-parametric, for example models similar to \cite{xu06:infinite,kemp06:learning}, and using the ideas presented in \cite{lloyd12:random}.
We also want to investigate the utility of other inference techniques, such as Gibbs sampling and variational Bayes methods.
Further work on evaluation of the approach on more real-world datasets is also of interest.

\subsubsection*{Acknowledgments}
The authors would like to thank Lucas Bordeaux, Andy Gordon, Tom Minka, and John Guiver for valuable discussions and insights.
We are also grateful for the feedback from the anonymous reviewers of the NIPS 2012 workshop on probabilistic programming.

\bibliographystyle{unsrtnat}
\bibliography{../sameer}
\end{document}